\documentclass[11pt,a4paper]{article}
\usepackage[hyperref]{emnlp2020}
\usepackage{times}
\usepackage{latexsym}

\usepackage{microtype}
\usepackage{amsmath}
\usepackage{graphicx}
\usepackage{makecell}
\usepackage{array}
\usepackage{multirow}
\usepackage{xcolor}

\usepackage{multirow}

\aclfinalcopy 
\def\aclpaperid{1830} 

\title{Reducing Quantity Hallucinations in Abstractive Summarization}

\author{Zheng Zhao \quad Shay B. Cohen \quad Bonnie Webber \\
  Institute for Language, Cognition and Computation \\
  School of Informatics, University of Edinburgh \\
  10 Crichton Street, Edinburgh, EH8 9AB \\
  \texttt{zheng.zhao@ed.ac.uk} ,
  \texttt{\{scohen,bonnie\}@inf.ed.ac.uk}}

\date{}

\begin{document}
\maketitle
\begin{abstract}
It is well-known that abstractive summaries are subject to hallucination---including material that is not supported by the original text. While summaries can be made hallucination-free by limiting them to general phrases, such summaries would fail to be very informative. Alternatively, one can try to avoid hallucinations by verifying that any specific entities in the summary appear in the original text in a similar context. This is the approach taken by our system, \textsc{Herman}. The system learns to recognize and verify quantity entities (dates, numbers, sums of money, etc.) in a beam-worth of abstractive summaries produced by state-of-the-art models, in order to up-rank those summaries whose quantity terms are supported by the original text. Experimental results demonstrate that the ROUGE scores of such up-ranked summaries have a higher Precision than summaries that have not been up-ranked, without a comparable loss in Recall, resulting in higher F$_1$. Preliminary human evaluation of up-ranked vs. original summaries shows people's preference for the former.
\end{abstract}

\section{Introduction}

Automatic summarization is the task of compressing a lengthy text to a more concise version that preserves the information of the original text. Common approaches are either \textit{extractive}, selecting and assembling salient words, phrases and sentences from the source text to form the summary \citep{lin-bilmes-2011-class,DBLP:conf/aaai/NallapatiZZ17,narayan-etal-2018-ranking}, or \textit{abstractive}, generating the summary from scratch, containing novel words and phrases that are paraphrased from important parts of the original text \citep{lapata-2005, rush-etal-2015-neural, wang-etal-2019-biset}. The latter is more challenging as it involves human-like capabilities, e.g., paraphrasing, generalizing, inferring and including  real-world knowledge \citep{see-etal-2017-get}. 

\begin{table}[t]
\small
\centering
\begin{tabular}{m{7.2cm}}
\Xhline{1pt}
\textbf{Article:} \dots the volcano was {\color{cyan}still spewing ash on Sunday}, hampering rescue operations. {\color{cyan}More than a dozen people were killed when it erupted in 2014} \ldots rescue teams are still scouring the area, looking for {\color{cyan}more victims} who may have been killed or badly burned \ldots\\ 
\textbf{Summary:} Rescue teams in Indonesia are searching for {\color{red}more than 20 people} missing after the Mount Sinabung volcano erupted on {\color{red}Saturday}, killing {\color{red}at least 11 people} and injuring {\color{red}at least 20 others}.\\
\hline
\textbf{Article:} The scale of the criminal operation has been detailed by {\color{cyan} the three sources,} who say they were \ldots a victim of the fraud shown the call centre script has confirmed it matched the one read out to her when {\color{cyan}she was conned out of £5,000} \ldots\\ 
\textbf{Summary:} {\color{green}Three whistleblowers} have told the BBC that they were involved in a scam that conned hundreds of TalkTalk customers out of {\color{red}more than £100,000}.\\
\hline
\textbf{Article:} The government and the doctors' union have agreed to continue negotiating {\color{cyan}until Wednesday}.
The talks, hosted by conciliation service Acas \ldots\\ 
\textbf{Summary:} Talks aimed at averting the imposition of a new junior doctors' contract in England have been extended for {\color{red}a second day}.\\
\Xhline{1pt}
\end{tabular}
\caption{\label{tab:hallucination-example} Examples of system generated abstractive summaries with hallucinated quantities. Phrases in the articles highlighted in {\color{cyan}cyan} have been used by the summarization system to generate summaries. Phrases in the summaries highlighted in {\color{green}green} are correct with respect to the article, whereas {\color{red}red} highlighting indicates hallucinations. Note that the first article describes both a new eruption and a previous one in 2014. It was in the previous eruption that \textit{more than a dozen people} were killed, hence a hallucination of \textit{at least 11 people} killed and \textit{at least 20} injured in the \textbf{new} eruption.}
\end{table}

Abstractive summarization has attracted increasing attention recently, thanks to the availability of large-scale datasets \citep{sandhaus2008new, DBLP:conf/nips/HermannKGEKSB15, DBLP:conf/naacl/GruskyNA18, narayan-etal-2018-dont} and advances on neural architectures \citep{DBLP:conf/nips/SutskeverVL14,DBLP:journals/corr/BahdanauCB14,DBLP:conf/nips/VinyalsFJ15,NIPS2017_7181}. Although modern abstractive summarization systems generate relatively fluent summaries, recent work has called attention to the problem they have with factual inconsistency \citep{kryscinski-etal-2019-neural}. That is, they produce summaries that contain hallucinated facts that are not supported by the source text. A recent study has shown that up to 30\% of summaries generated by abstractive summarization systems contain hallucinated facts \citep{DBLP:conf/aaai/CaoWLL18}. Such high levels of factual hallucination raise serious concern about the usefulness of abstractive summarization, especially if one believes that summaries (whether extractive or abstractive) should contain a mixture of general and specific information \citep{louis-nenkova-2011-text}.

This paper explores reducing the frequency of one type of hallucinated fact in abstractive summaries---\textit{hallucinated quantities}. We focus on quantities not only because they are important for factual consistency, but also because, unless they are wildly inaccurate, a reader might not notice that they are hallucinated. Moreover, unlike people's names (which are also frequently hallucinated), quantity entities are rarely referred to anaphorically, avoiding the need to resolve anaphoric expressions, making them an excellent testbed for the study of hallucination. The quantities we address can be broadly categorized into seven types: dates, times, percentages, monetary values, measurements, ordinals, and cardinal numbers. Table \ref{tab:hallucination-example} shows some examples of hallucinated quantities introduced by abstractive summarization models. 

We present \textsc{Herman}\footnote{Name inspired by the fact-checker Herman Brooks from the 1980s American sitcom ``Herman's Head.''}, a system that learns to recognize quantities in a summary and verify their factual consistency with the source text. Our system can be easily coupled with any abstractive summarization models that produce a beam-worth of candidate summaries. After verifying consistency, we use a re-ranking approach that up-rank those summaries whose quantities are supported by the source text, similar to the method proposed by \citet{falke-etal-2019-ranking}. Training data is automatically generated in a weakly supervised manner from a summarization dataset containing both original and synthetic data. The synthetic data is created by selecting quantity entities from the summary and replacing them with randomly selected entities from the source text that are the same type. We perform experiments on the XSum dataset \citep{narayan-etal-2018-dont} which favors an abstractive modeling approach. Results based on automatic evaluation using ROUGE \cite{lin-2004-rouge} demonstrate that up-ranked summaries have higher ROUGE Precision than original summaries produced by three different summarization systems. While ROUGE Recall of these up-ranked summaries is lower, overall ROUGE F$_1$ is higher for up-ranked summaries, showing that it is not simply a like-for-like trade-off of Recall for Precision. A preliminary human evaluation study shows that subjects prefer the up-ranked summaries to the original summaries. 

\begin{table*}[t]
\centering
\begin{tabular}{lccccccccccc}
\Xhline{1pt}
{\textbf{Article}} & \multicolumn{11}{p{13.3cm}}{The crash happened at Evanton at about 17:20 on Saturday. The fire service and the air ambulance was sent to the scene.
{ \color{cyan} The occupants of all three vehicles were injured,} but the extent of their injuries was not known, police said.
A spokesman added: ``Inquiries are ongoing into this matter and no further witnesses are sought at this time'' \ldots}\\ 
\textbf{Summary} & {\color{green} Several} & people & have & been & injured & in & a & {\color{green}three-car} & collision & on & \ldots\\
\textbf{$Y$ labels} & \texttt{\color{green}B-V} & \texttt{O} & \texttt{O} & \texttt{O} & \texttt{O} & \texttt{O} & \texttt{O} & \texttt{\color{green}B-V} & \texttt{O} & \texttt{O} & \ldots\\
\textbf{$M$ labels} & \texttt{\color{green}1} & \texttt{0} & \texttt{0} & \texttt{0} & \texttt{0} & \texttt{0} & \texttt{0} & \texttt{\color{green}1} & \texttt{0} & \texttt{0} & \ldots\\
\textbf{$z$ label} & \multicolumn{11}{c}{\texttt{VERIFIED}}\\
\Xhline{1pt}
\end{tabular}
\caption{\label{tab:BIO-example} An example of a \texttt{VERIFIED} summary with its labels from our dataset. 
 {\color{cyan}Cyan} text highlights the support
in the source document for the quantity token highlighted {\color{green}green} in the summary.}
\end{table*}

\section{Related Work}
Recent studies have suggested that abstractive summarization systems are prone to generate summaries with hallucinated facts that cannot be supported by the source document. \citet{DBLP:conf/aaai/CaoWLL18} reported that almost 30\% of the outputs of a state-of-the-art system contain factual inconsistencies. An evaluation of summaries produced by recent state-of-the-art models via crowdsourcing suggested that 25\% of the summaries have factual errors \citep{falke-etal-2019-ranking}. The work also showed that ROUGE scores do not correlate with factual correctness, emphasizing that ROUGE based evaluation alone is not enough for summarization task. In addition, \citet{kryscinski-etal-2019-neural} pointed out that current evaluation protocols correlate weakly with human judgements and do not take factual correctness into account. \citet{DBLP:journals/corr/abs-2005-00661} conducted a large scale human evaluation on the generated summaries of various abstractive summarization systems and found substantial amounts of hallucinated content in those summaries. They also concluded that summarization models initialized with pre-trained parameters perform best on not only ROUGE, but also human judgements of faithfulness/factuality.

Another line of research focused on evaluating factual consistency of summarization systems. \citet{DBLP:journals/corr/abs-1910-12840} proposed a weakly-supervised, model-based approach for evaluating factual consistency between source documents and generated summaries. They first generate training data by applying a series of transformations to randomly selected individual sentences from source documents (which they call \textit{claims}) and assign them a binary label based on the type of the transformation. Then they train a fact-checking model to classify the label of the claim and extract spans in both the source document and the generated summary explaining the model's decision. \citet{goodrich-2019-assess} introduced a model-based metric for estimating the factual accuracy of generated text. Factual accuracy is defined as Precision between claims made in the source document and the generated summary, where claims are represented as \textit{subject-relation-object} triplets. They also released a new dataset for training relation classifiers and end-to-end fact extraction models based on Wikipedia and Wikidata.

Several studies have focused on tackling the problem of factual inconsistencies between inputs and outputs of summarization models by exploring different model architectures and methods for training and inference. \citet{DBLP:conf/aaai/CaoWLL18} attempted to solve the problem by encoding extracted facts as additional inputs to the system. The fact descriptions are obtained by leveraging Open Information Extraction \citep{DBLP:conf/ijcai/BankoCSBE07} along with parsed dependency trees of the input text. \citet{DBLP:journals/corr/abs-1911-02541} developed a framework to evaluate the factual correctness of generated summaries by employing an information extraction module to check facts against the source document, and proposed a training strategy that optimizes the model using reinforcement learning with factual correctness as a reward policy. \citet{falke-etal-2019-ranking} proposed a re-ranking approach to improve factual consistency of summarization models. Their approach used natural language inference (NLI; \citealt{bowman-etal-2015-large}) models to score candidate summaries obtained in beam search by averaging the entailment probability between all sentence pairs of source document and summary. The summary with the highest score is up-ranked and used as final output of the summarization system. After evaluating their approach using summaries generated by summarization systems trained on the CNN-DailyMail corpus \citep{DBLP:conf/nips/HermannKGEKSB15}, they concluded that out-of-the-box NLI models transfer poorly to the task of evaluating factual correctness, limiting the effectiveness of re-ranking.

{
\begin{figure*}[t]
\centering
\includegraphics[scale=0.90]{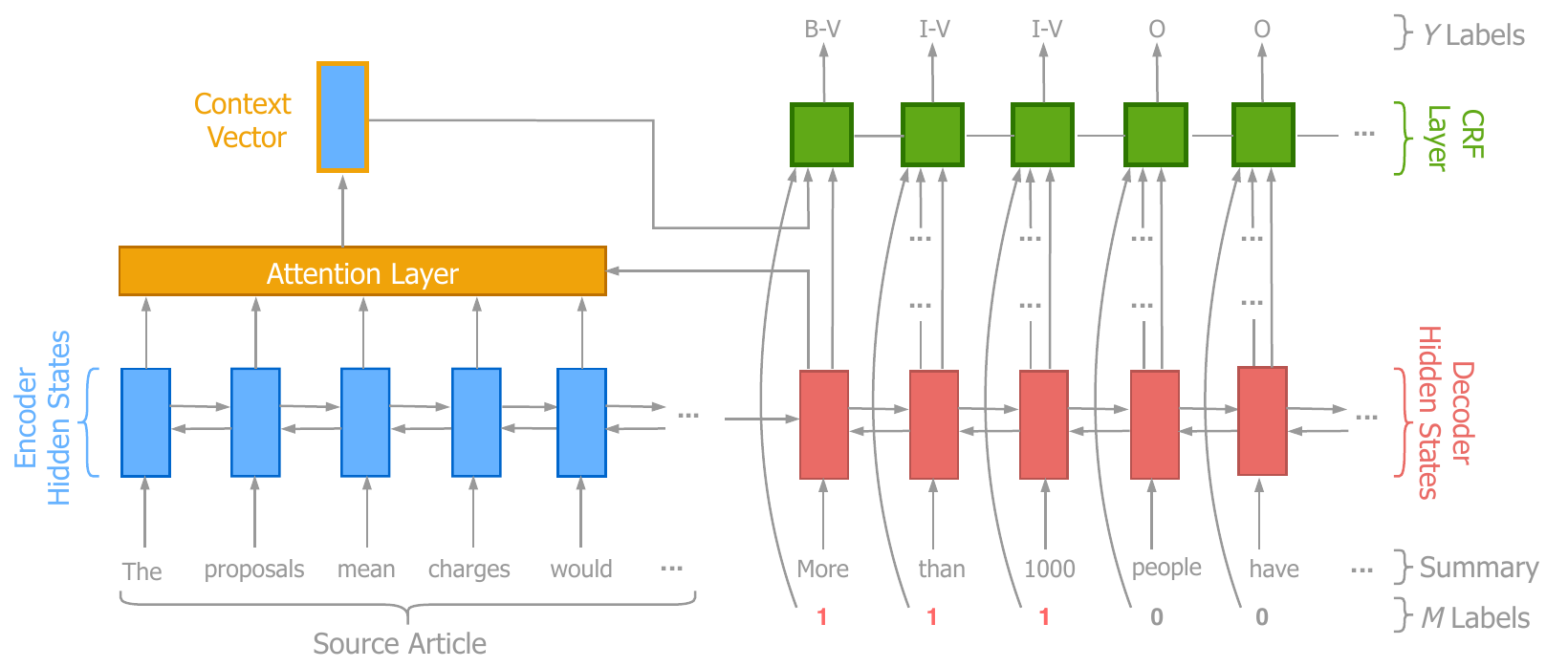}
\caption{\label{fig:model_architecture} Architecture of \textsc{Herman}. Note that the binary classifier for predicting whether a summary is verified ($z$ labels) is omitted here. It simply takes the context vectors of the summary tokens and run through a MLP classifier. }
\end{figure*}
}

\section{Methodology}

Let $\mathit{X}$ be the article and $S$ be the corresponding summary where both are sequences of tokens, $x_1 \cdots x_a$ and $s_1 \cdots s_n$, respectively. Given a $(X, S)$ pair, our aim is to generate a tag sequence $Y$ with the same length as $S$ (i.e., $n$) and a summary-level label $z \in \{\texttt{VERIFIED},\texttt{UNVERIFIED}\}$, indicating whether the summary $S$ can be verified using $X$. The generated tag sequence $y_1 \cdots y_n$ contains token-level labels where $y_j \in \{$\texttt{B-V}, \texttt{B-U}, \texttt{I-U}, \texttt{I-V}, \texttt{O}$\}$ indicating whether the token is Verified, Unverified, or Other. We adopt the BIO format \citep{ramshaw1999text} for labels since entities may span multiple tokens. To aid the recognition of quantity based entities, we also obtain a sequence of binary labels $ M = ( m_1,\ldots , m_n )$ for the summary indicating the location of these entities.  

Our approach consists of two steps. First, we create a synthetic, weakly-supervised dataset $\mathcal{D} = \{ (X^{(i)}, S^{(i)}, M^{(i)}, Y^{(i)}, z^{(i)}) \mid i \in \{ 1\ldots N \} \}$ consisting of $N$ input-output pairs, where $X$, $S$, and $M$ are the input, $Y$ and $z$ are the output. At training time, a \emph{verification model} learns to recognize and verify quantities in the summary. At test time, the same verification model is applied to the summaries identified in a beam search for candidate summaries carried out by the summarization systems, which results in each of them being given a verification score. We provide a detailed description in the rest of this section.

\subsection{Dataset Generation}
\label{ssec:dataset_generation}
The dataset used to train the verification model comprises the dataset used to train the summarization system, augmented with negative examples and additional labels. As we focus on quantities, we apply the spaCy NER tagger \citep{spacy2} to identify all such entities in both the article and summary. A gold summary in the original summarization dataset receives a $z$ label \texttt{VERIFIED}. To generate versions of this summary with $z$ label \texttt{UNVERIFIED}, we replace quantity entities in the summary with randomly selected entities from the article that are the same type. For example, a date entity can only be replaced by another date entity from the article. We ensure the \texttt{UNVERIFIED} summary is different from its  \texttt{VERIFIED} counterpart. If an article only contains the one quantity entity which appears in the \texttt{VERIFIED} summary, i.e. no replacement can be found to get the \texttt{UNVERIFIED} version, we discard both examples for our dataset to maintain a balanced dataset.

In addition to the binary summary-level label $z$, we also generate two sequences of labels $Y$ and $M$. Quantity entities recognized by spaCy NER in \texttt{VERIFIED} summaries are labeled \texttt{V}, and replaced ones in the \texttt{UNVERIFIED} summaries are labelled \texttt{U}. Tokens with \texttt{O} labels are unlikely to directly affect whether a quantity based entity has been hallucinated, whereas tokens with \texttt{V} and \texttt{U} labels indicate they are important and could potentially affect the factual accuracy of the summary. With BIO format adopted, these labels become \texttt{B-V}, \texttt{B-U}, \texttt{I-V}, \texttt{I-U}, and \texttt{O}. For the sequence of binary labels $M$:

{
\[
    m_j= 
\begin{cases}
    0,& \text{if } y_j = \texttt{O}\\
    1,              & \text{otherwise}
\end{cases}.
\]
}

\noindent Table \ref{tab:BIO-example} illustrates an example of \texttt{VERIFIED} summary with its labels and corresponding article.

\subsection{Verification Model}
The overall architecture for our verification model \textsc{Herman} is illustrated in Figure \ref{fig:model_architecture}. The article encoder provides hidden representations for every input token which are then fed to a decoder with attention to obtain the context vector. The context vectors from every token in the summary are then fed into a Conditional Random Fields (CRF) layer \citep{LaffertyMP01} to generate the tag sequence $Y$. The same context vectors are fed into a binary classifier to obtain the binary label $z$.

\paragraph{BiLSTM Article Encoder} For input article $X$ where $X = \{x_1, \ldots,x_a\}$ and $x_i$ denotes the $i$th token in $X$, a contextualized token-level encoding $h_i$ is obtained via a BiLSTM encoder \citep{hochreiter1997long}:
{
\[ \overrightarrow{h}_i = \mathbf{LSTM}_f(x_i, \overrightarrow{h}_{i-1}),\]
\[ \overleftarrow{h}_i = \mathbf{LSTM}_b(x_i, \overleftarrow{h}_{i+1}),\]
\[ h_i = [\overrightarrow{h}_i;\overleftarrow{h}_i],\]
}
\noindent where $\overrightarrow{h}_i$ and $\overleftarrow{h}_i$ are hidden states of forward and backward LSTMs at time step $i$, and $;$ denotes the concatenation operation. 

\paragraph{BiLSTM-CRF Decoder with Attention}
The decoder generates sequence of labels $Y$ as well as a binary label $z$. As the length of labels to be decoded is fixed, the setup is similar to BiLSTM-CRF used in the sequence tagging task \citep{Huang2015BidirectionalLM}. The difference is that the decoder takes additional input $h_i$ which is article encoding and incorporates attention mechanism \citep{BahdanauCB14}. The BiLSTM with attention component first encodes the summary, token by token, to produce an intermediate representation. We also obtain a sequence of binary labels $ M = \{m_1,\ldots , m_n\}$ for the summary using spaCy NER to recognize tokens that make up quantity entities. Then the intermediate representation, along with the binary label sequence, is fed to the CRF layer to predict the $Y$ label. The intermediate representation is also fed to an MLP classifier to obtain the binary label $z$.

\subsection{Training and Inference}
Given the training set with labelled sequence $\{ X^{(i)}, S^{(i)}, M^{(i)}, Y^{(i)}, z^{(i)} \mid i \in \{ 1\ldots N \} \}$, we maximize the conditional log likelihood for the local verification objective:
\[ \bar{w} = \mathop{\mathrm{argmax}}_w \sum_{i=1}^{N} \log p(Y^{(i)} \mid X^{(i)}, S^{(i)}, M^{(i)}, w),\]
\noindent where $w$ denotes the model's parameters including the weights of the LSTMs and the transition weights of the CRF. The loss function for $Y$ labels is the negative log-likelihood based on $Y^{(i)} = \{y_1, \ldots, y_n\}$: 
\[ \mathcal{L}_\mathit{Y} = - \sum_{i=1}^{N} \sum_{j=1}^{n} \log p(y_j),\]

\noindent where $y_j\in Y^{(i)}$. For global verification which is predicting $z$ label, the loss function is the binary cross entropy:

\begin{multline*}
    \mathcal{L}_\mathit{z} = \sum_{i=1}^{N} z^{(i)} \log p(z^{(i)}) \\+ (1 - z^{(i)}) \log (1 - p(z^{(i)})).
\end{multline*}

The final objective which combines both local and global verification is defined as the following:

\[ \mathcal{L} = \alpha \mathcal{L}_\mathit{Y} + (1 - \alpha) \mathcal{L}_\mathit{z},\]

\noindent where $\alpha \in [0,1]$ is a hyperparameter indicating weight balance between $\mathcal{L}_\mathit{Y}$ and $\mathcal{L}_\mathit{z}$. At test time, inference for a summary $S$ is obtained by applying Viterbi algorithm at the CRF layer to find the most probable sequence $\hat{Y}$: 

\[ \hat{Y} = \mathop{\mathrm{argmax}}_Y P(Y \mid X, S, M, \bar{w}).\]

\subsection{Re-ranking to Avoid Hallucination}
\label{sec:rerank}
We adopt a re-ranking approach in order to reduce the frequency of hallucinated quantities in the output of abstractive summarization. This is similar to the approach taken by \citet{falke-etal-2019-ranking} with the difference being that their system's inputs are sentence level whereas ours are document-level. Assume an abstractive summarization system can produce a list of $k$ candidate summaries $S_1,\ldots,S_k$ for a given document $X$ using beam search, we leverage predictions of \textsc{Herman} to give each summary a verification score. Our scoring approach has two variants: \textsc{Herman-Global}, and \textsc{Herman-Local}. \textsc{Herman-Global} uses the raw output of global verification label $z$ which has a real value between $[0,1]$. \textsc{Herman-Local} uses the average probabilities of \texttt{B-V}, \texttt{B-U}, \texttt{I-V}, and \texttt{I-U} labels where entries of \texttt{B-U} and \texttt{I-U} are counted negatively. Out of the $k$ candidate summaries, the summary with the highest verification score is selected as the final generated summary for the summarization system.

\section{Dataset}
\label{sec:Dataset}

\begin{table*}[t]
\centering
\begin{tabular}{ccccccc}
\Xhline{1pt}
\textit{Date} & \textit{Time} & \textit{Percent} & \textit{Money} & \textit{Quantity} & \textit{Ordinal} & \textit{Cardinal}\\
29\% & 2\% & 1\% & 4\% & 1\% & 8\% & 25\% \\
\Xhline{1pt}
\end{tabular}
\caption{\label{tab:quantity_entity_dist_gold}
The distribution of quantity entities in the XSum dataset. Note that the percentages sum to more than 55\%, as a summary can contain more than one type of quantity entity. For more details regarding the types of entities, please refer to the official spaCy webpage\footnotemark.
}
\end{table*}

We use the XSum dataset which was developed for abstractive document summarization \citep{narayan-etal-2018-dont}. The XSum dataset consists of BBC articles, with a single-sentence summary of each. This summary is a professionally written introductory sentence, typically written by the author of the article, which is separated from the article, with the remaining text taken to be the document. This one-sentence summary, different from a headline whose purpose is to attract readers to read the article, draws on information distributed in various parts of the document and displays multiple levels of abstraction including paraphrasing, fusion, synthesis, and inference. The dataset contains 204,045 instances for training, 11,332 instances for validation, and 11,334 instances for testing. Overall, 55\% of the instances contain at least one quantity. The distribution of quantity entities is shown in Table~\ref{tab:quantity_entity_dist_gold}. It is clear that the different types of quantities are distributed unevenly: While almost 30\% of summaries contain at least one \textit{date} entity, only 1\% contain at least one \textit{quantity} entity. Due to the way in which the summary was created for a document, the summary often contains phrases that do not appear in the document itself. In fact, fewer than 16\% of the summaries in the test set have quantity tokens that also appear in their corresponding documents.

In order to obtain the dataset used to train \textsc{Herman}, we follow procedures described in Section~\ref{ssec:dataset_generation}. We apply same pre-processing steps noted by \citet{narayan-etal-2018-dont}. We also truncate the input document to 400 tokens and limit the length of the summary to 90 tokens. The dataset size for training, validation, and test are 190,370, 10,594, and 10,592, respectively.
As noted in Section~\ref{ssec:dataset_generation}, the dataset we use is smaller than the XSum dataset because we discard instances which cannot be perturbed to obtain an \texttt{UNVERIFIED} summary.\footnotetext{{\url{https://spacy.io/api/annotation\#named-entities}}}

\section{Experiments}
\label{sec:experiment}
For all experiments, we set the hidden dimensions to 256, the word embeddings to 100, and the vocabulary size to 50k. The word embeddings are initialized using pre-trained GloVe \citep{pennington2014glove} vectors (6B tokens, \textit{uncased}). We also experimented using a pre-trained, \textit{base-uncased} BERT \citep{devlin-etal-2019-bert} for word embedding initialization. Our training used the Adam optimizer \citep{KingmaB14} with a learning rate of 0.001. We also use gradient clipping with a maximum gradient norm of 5 and we do not use any kind of regularization. We use loss on the validation set to perform early stopping. We set $\alpha$ to 0.66, suggesting local verification is more important than global verification. Our model was trained on a single GeForce GTX 1080 Ti GPU with a batch size of 32. We use PyTorch \citep{NEURIPS2019_9015} for our model implementation. For CRF, we used the AllenNLP library \citep{DBLP:journals/corr/abs-1803-07640} with constrained decoding for the BIO scheme. To evaluate our verification model, we need outputs from abstractive summarization systems. We obtain those from three selected systems: \textsc{TConvS2S} \citep{narayan-etal-2018-dont}, \textsc{BertSum} \citep{liu-lapata-2019-text}, and BART \citep{bart-abs-1910-13461} using pre-trained checkpoints provided by the authors.

\begin{table}
\centering
\begin{tabular}{cccc}
\Xhline{1pt}
{Label} & {Precision} & {Recall} & {F$_1$}\\
\Xhline{1pt}
\texttt{B-V} & 75.18 & 78.13 & 76.63\\
\texttt{B-U} & 75.11 & 71.28 & 73.14\\
\texttt{I-V} & 84.78 & 85.63 & 85.20\\
\texttt{I-U} & 83.86 & 83.93 & 83.89\\
\texttt{O} & 100.0 & 100.0 & 100.0\\
\Xhline{1pt}
\end{tabular}
\caption{\label{tab:verification_result}
Results of \textsc{Herman} on the test set using GloVe word embedding. 
}
\end{table}

\begin{table}
\centering
\begin{tabular}{cccc}
\Xhline{1pt}
{Label} & {Precision} & {Recall} & {F$_1$}\\
\Xhline{1pt}
\texttt{B-V} & 72.83 & 81.24 & 76.81\\
\texttt{B-U} & 75.73 & 69.28 & 72.37\\
\texttt{I-V} & 84.58 & 87.27 & 85.90\\
\texttt{I-U} & 85.03 & 83.47 & 84.24\\
\texttt{O} & 100.0 & 100.0 & 100.0\\
\Xhline{1pt}
\end{tabular}
\caption{\label{tab:verification_result_bert}
Results of \textsc{Herman} on the test set using BERT word embedding. 
}
\end{table}

\section{Results}

\paragraph{Automatic Evaluation}
We first present results in Table~\ref{tab:verification_result} from our verification model using GloVe on the test set. On the binary classification task of determining whether a summary is \texttt{VERIFIED} or \texttt{UNVERIFIED}, the model achieved accuracy of 80.12 and F$_1$ of 80.94. The results using BERT are displayed in Table \ref{tab:verification_result_bert}. The model attained accuracy of 80.23 and F$_1$ of 81.6. While no significant difference can be observed in performance, using BERT does triple the needed training time, so does not seem justified.

\begin{table*}[t]
\centering
{\small
\begin{tabular}{|l|l|ccc|ccc|ccc|c|}
\Xhline{1pt}
 & {Model} & {R1-R} & {R1-P} & {R1-F}
        & {R2-R} & {R2-P} & {R2-F}
        & {RL-R} & {RL-P} & {RL-F} & {avg-Q}\\
\Xhline{1pt}
\multirow{5}{*}{\rotatebox{90}{\textsc{BART}}} & Baseline-shortest & 45.50 &  \textbf{46.95} &  \textbf{45.40}  & 21.86   & \textbf{22.61} &  \textbf{21.83}  & 36.80 &  \textbf{38.01}  & \textbf{36.74} & 0.69
\\
& Baseline-max-overlap & 49.46 &  41.66 &  44.55 &  23.35 &  19.57  & 20.97  & 39.30 &  33.08  & 35.38 & \textbf{0.95}
\\
& Original & \textbf{49.64} &  41.54 &  44.57 &  \textbf{23.43}  & 19.50 &  20.96 &  \textbf{39.39} &  32.95  & 35.36 & 0.89
\\
& \textsc{Herman-Local} & 48.51 &  42.78 &  44.73 &  22.97 &  20.20  & 21.14 &  38.70 &  34.12  & 35.68 & 0.88
\\
& {\textsc{Herman-Global}} & 47.88 &  {43.52} &  {44.79} &  22.66 &  {20.56}  & {21.17}  & 38.26 &  {34.79} &  {35.80} & 0.92
\\

\hline
\multirow{5}{*}{\rotatebox{90}{{\textsc{BertSum}}}} & Baseline-shortest & 36.78 &  \textbf{42.26} & 38.71  & 15.61 &  \textbf{17.87} &  16.38  & 29.71 &  \textbf{33.91} &  31.16 & 0.62
 \\
& Baseline-max-overlap & 38.17 &  41.25  & 39.01 &  \textbf{16.28}  & 17.50 &  16.58 &  30.66  & 32.94 &  31.24 & 0.76
\\
& Original & 38.37 &  40.73 &  38.86 &  16.24  & 17.13 &  16.38 &  \textbf{30.75}  & 32.44  & 31.04 & 0.65 
\\
& \textsc{Herman-Local} & \textbf{38.45} &  40.14  & 38.63  & 16.12 &  16.72  & 16.12  & 30.71 &  31.87 &  30.75 & 0.79
\\
& \textsc{Herman-Global} & 37.99 &  41.59  & \textbf{39.06}  & 16.24 &  17.70 &  \textbf{16.65} &  30.59 &  33.28  & \textbf{31.36} & \textbf{0.81}
\\
\hline
\multirow{5}{*}{\rotatebox{90}{{\textsc{TConvS2S}}}} & Baseline-shortest & 27.43 &  \textbf{37.28}  & 30.99  & 9.84  &  \textbf{13.49} &  11.15 &  22.43  & \textbf{30.41} &  25.32 & 0.45
\\
& Baseline-max-overlap & 30.19  & 34.57 &  31.64 &  10.79  & 12.34  & 11.29  & 24.37  & 27.81 &  25.50 & 0.71
\\
& Original & \textbf{30.42} &  34.63  & 31.80  & 10.96 &  12.46  & 11.45  & \textbf{24.58} &  27.89  & 25.66 & 0.58
 \\
& {\textsc{Herman-Local}} & 29.95  & 34.50  & 31.43  & 10.59  & 12.16  & 11.09  & 24.17 &  27.72  & 25.31 & 0.75
\\
& {\textsc{Herman-Global}} & 30.36  & 34.82  & \textbf{31.85} &  \textbf{10.98} &  12.59  & \textbf{11.51} &  24.56 &  {28.08}  & \textbf{25.72} & \textbf{0.78}
\\
\Xhline{1pt}
\end{tabular}

}
\caption{\label{tab:rouge_result}
Automatic evaluation on the XSum test set. Each of the three horizontal sections reports scores for one of the three abstractive summarization systems: BART, \textsc{BertSum} and \textsc{TConvS2S}. For each system, we present ROUGE scores for the two baseline models, the one original model, and the two variants of our \textsc{Herman} model.
Baseline-shortest refers to the model that selects the shortest summary. 
Baseline-max-overlap refers to the model that selects the summary which overlaps the most with the source document in terms of quantity entities . avg-Q denotes the average number of quantity entities per summary.  
}
\end{table*}

The standard automatic evaluation metric for summarization is ROUGE. We report the Precision, Recall and F$_1$ scores of ROUGE-1/2/L, which respectively measure the word-overlap, bigram-overlap, and longest common sequence between system and reference summaries. Using \textsc{Herman}, we obtain verification scores for the full beam of candidate summaries produced by the summarization systems. We re-rank candidate summaries using the verification score as described in Section \ref{sec:rerank} and evaluate the up-ranked summaries. In addition to \textsc{Herman-Global} and \textsc{Herman-Local}, we also introduce two baseline re-ranking approaches: the first selects the shortest summary from the beam, and the second selects the summary with maximum quantity entity overlap with the source document. The results on the XSum dataset are shown in Table \ref{tab:rouge_result}. While selecting the shortest summary is a very strong baseline, outperforming all other systems in ROUGE-1/2/L Precision, we can still see that \textsc{Herman-Global} has the best performance in ROUGE-1/2/L Precision and F$_1$ despite that baseline. After re-ranking by \textsc{Herman-Global}, 17.27\% originally ranked top summaries produced by BART stayed at the top rank. While \textsc{BertSum} had nearly the same, only 9.05\% of the summaries produced by \textsc{TConvS2S} stayed top-ranked, so if re-ranking leads to improvements, it would be even more helpful in the case of \textsc{TConvS2S}.

The first thing to note is that the up-ranked summaries have a lower ROUGE Recall than other models. This is common with any model that filters output, since it can exclude items that might otherwise contribute to Recall. 
ROUGE-1/2/L Precision increases after re-ranking as the verification model ensures summaries with more verified content will be ranked higher in the beam. More verified content also means more tokens appearing in the document and reference summary. Overall, ROUGE-1/2/L F$_1$ score for up-ranked summaries exceeds that of original summaries. To analyze the effect of our systems on quantity entities, we also compute average number of quantity entities per summary for each system. The baseline that selects the summary with maximum quantity entity overlap with the source document, not surprisingly, has very high averages and achieved the highest number for BART. \textsc{Herman-Global} achieves highest average for \textsc{BertSum} and \textsc{TConvS2S}. In BART, it follows the baseline closely at second place. Together with its ROUGE performance, this indicates that our model not only encourages the inclusion of quantity entities in the summary, but also includes them correctly.

\begin{table*}
\small
\centering
\begin{tabular}{|l|rrr|rrr|rrr|}
\Xhline{1pt}
\multirow{1}{*}{Quantity} & \multicolumn{3}{c|}{{\textsc{BART}}}& \multicolumn{3}{c|}{{\textsc{BertSum}}}& \multicolumn{3}{c|}{{\textsc{TConvS2S}}}\\
\multicolumn{1}{|c|}{Type} & Original & Up-ranked & \% diff & Original & Up-ranked & \% diff & Original & Up-ranked & \% diff \\
\Xhline{1pt}
\textit{Date} & 3,865 & 4,284 & 11\% & 2,735 & 3,731 & 36\% & 2,347 & 3,747 & 60\% \\
\textit{Time} & 208 & 221 & 6\% & 92 & 160 & 74\% & 47 & 75 & 60\% \\
\textit{Percent} & 119 & 118 & -1\% & 96 & 93 & -3\% & 93 & 102 & 10\% \\
\textit{Money} & 177 & 166 & -6\% & 450 & 545 & 21\% & 291 & 379 & 30\% \\
\textit{Quantity} & 45 & 37 & -18\% & 41 & 45 & 10\% & 17 & 16 & -6\% \\
\textit{Ordinal} & 1,330 & 1,194 & -10\% & 959 & 1,057 & 10\% & 1,200 & 1,208 & 1\% \\
\textit{Cardinal} & 2,924 & 2,940 & 1\% & 2,327 & 2,580 & 11\% & 2,048 & 2,418 & 18\% \\
\hline
{All Types} & 6,612 & 6,835 & 3\% & 5,486 & 6,405 & 17\% & 4,905 & 6,192 & 26\%\\
\Xhline{1pt}
\end{tabular}
\caption{\label{tab:quantity_entity_dist_testset}
The statistics of different types of quantity entities on test set summaries for all three abstractive summarization systems: BART, \textsc{BertSum} and \textsc{TConvS2S}. For each system, we provide the number of original summaries and up-ranked summaries that contain at least one instance of the given type of quantity entity. Up-ranked summaries are produced by \textsc{Herman-Global}. \% diff denotes the percentage difference between the number of up-ranked summaries and the number of original summaries for a given quantity type. 
}
\end{table*}

To further analyze how our approach affects the distribution of different types of quantity entities, we also computed test set statistics for both original summaries produced by the summarization systems and up-ranked summaries produced by \textsc{Herman-Global}. The results are provided in Table~\ref{tab:quantity_entity_dist_testset}. Overall, counting all quantity types, we can see that BART encourages the inclusion of quantities the most, for both original and up-ranked summaries, while \textsc{TConvS2S} has the fewest summaries with quantity entities. However, the number of up-ranked summaries that contain at least one quantity increases the most for \textsc{TConvS2S}, a 26\% increase compared with the original summaries. This agrees with our prior point that as \textsc{TConvS2S} has the fewest summaries that remained top after re-ranking, our approach should be most helpful for \textsc{TConvS2S}. Looking at individual quantity types, the number of summaries containing \textit{date} or \textit{time} quantities increases across-the-board through re-ranking. For \textsc{BertSum} and \textsc{TConvS2S}, re-ranking generally increases the number of summaries that contain a specific quantity type, with the exception of \textit{percent} in \textsc{BertSum} and \textit{quantity} in \textsc{TConvS2S} where they decreased slightly. We suspect the reason to be that these types are underrepresented in the dataset: Thus, there is insufficient data for the model to learn from. On the other hand, re-ranking in BART leads to more decreases of the number of summaries that contain a specific quantity type. The reason could be that BART already has the highest number of summaries that contain a specific quantity type before re-ranking, and quantity types with a decrease after re-ranking are generally underrepresented types like \textit{percent} and \textit{quantity}. Representative types like \textit{date} and \textit{cardinal} are still increased through re-ranking.

\paragraph{Human Evaluation}
\citet{falke-etal-2019-ranking} have argued convincingly that ROUGE is inadequate as a measure of hallucination and factual correctness. As such, we have begun to carry out human evaluation. We noted in Section~\ref{sec:Dataset} that the XSum reference summary may not be an accurate representation of the source article, in that less than 16\% of the test set reference summaries have quantity tokens that also appear in their corresponding articles. As a result, our human evaluation presented subjects with a text consisting of both the reference summary and the source article, to give subjects a full sense of its contents.

Subjects assessed 40 trials, each consisting of a text followed by two candidate summaries---the original summary produced by the summarization model and the up-ranked summary selected by \textsc{Herman-Global}. These two summaries also satisfied the condition of being very similar except for one quantity entity. 
The trials comprised 37 randomly selected text-summary pairs that satisfied the additional condition, plus three simple
\textit{catch trials} in which one of the candidate summaries has obvious hallucinated quantities that are never present in the source article, to check whether subjects were paying attention and following the instructions.
The order of the trials was randomized for each subject.  

\begin{table*}[h!]
\small
\centering
\begin{tabular}{m{7.5cm}|m{7.5cm}}

\Xhline{1pt}
\textbf{Article 49:}Interest rates for savers have fallen to new record lows, after {\color{cyan}hundreds} of cuts in recent months and {\color{cyan}more than 1,000} in the past year \ldots In research carried out for the BBC, the rate-checking firm Savings Champion recorded {\color{cyan}1,440} savings rate cuts last year and {\color{cyan}more than 230} so far \ldots 
& 
\textbf{Article 4:} A man has been charged with causing the death of a {\color{cyan}three-year-old} girl by dangerous driving in a crash involving eight vehicles. Thomas Hunter, {\color{cyan}58}, of Mansfield Road, Mansfield, was arrested after the crash on the A34 at Hinksey Hill, Oxford, on {\color{cyan}25 August} \ldots
\\
\textbf{Original Summary: } {\color{cyan}More than 1,500} savings rate cuts have been made by banks in the past year and {\color{cyan}more than 230} so far this year, the BBC has learned. 
& 
\textbf{Original Summary: }A man has been charged with causing the death of a {\color{cyan}three-year-old} girl by dangerous driving after a crash in which seven people were injured.
\\
\textbf{Up-ranked Summary: }
{\color{cyan}More than 1,000} savings rate cuts have been made by banks in the past year and {\color{cyan}more than 230} so far this year.
& 
\textbf{Up-ranked Summary: }A man has been charged with causing the death of a {\color{cyan}six-year-old} girl by dangerous driving after a crash in which seven people were injured.
\\
\hline
\textbf{Article 83: }Millions of people face a rise in their insurance bills this week-end, as a result of an increase in Insurance Premium Tax (IPT). From Sunday, IPT will increase from {\color{cyan}6\% to 9.5\%}, a rise that was announced by Chancellor George Osborne in his Summer Budget \ldots
&
\textbf{Article 24: } Shares in Paddy Power Betfair fell {\color{cyan}more than 5\%} despite the bookmaker reporting rising revenues and underlying profits \ldots But after the costs of last year's merger between Paddy Power and Betfair were taken into account the company reported a loss of £5.7m \ldots
\\
\textbf{Original Summary: }Car insurance premiums (IPT) will increase by {\color{cyan}9\%} from Sunday, the AA has said.
&
\textbf{Original Summary: }Shares in bookmaker Paddy Power Betfair fell {\color{cyan}6\%} after the company reported a loss for the final three months of last year.
\\
\textbf{Up-ranked Summary: }Car insurance premiums (IPT) will increase by {\color{cyan}9.5\%} from Sunday , the AA has announced. 
&
\textbf{Up-ranked Summary: }Shares in bookmaker Paddy Power Betfair fell {\color{cyan}7\%} after the company reported a loss for the final three months of 2016.
\\
\Xhline{1pt}
\end{tabular}

\caption{\label{tab:ex-all-agree} Example trials selected from our human evaluation. Quantity entities have been highlighted the same way we did for human evaluation. With article 49 and 83 (containing \textit{cardinal} and \textit{percentage} quantities), all subjects agree that the up-ranked summary is more faithful, while with article 4 and 24 (containing \textit{date} and \textit{percentage} quantities), all agree that the original summary is more faithful.}
\end{table*}

In presenting each trial, quantities in the summaries and those with the same type in the text were highlighted to make them easy to find. Subjects were asked to choose the one summary whose highlighted quantity entity is more faithful to the source article.
Subjects were also told not to select a summary based on any other factors such as its \textit{fluency} (i.e., Does the summary sound like well-formed English?). After subjects make a choice of summary, they are also asked whether they think both candidate summaries were equally faithful or equally unfaithful. We will show shortly how subjects can prefer one summary over the other, even while considering both to be faithful (or both to be unfaithful) to the original text.
This preliminary experiment was carried out on the Qualtrics platform, with three volunteer subjects. Each subject took between 35 and 45 minutes to finish.

While our results are still preliminary, they provide some evidence that subjects consider the up-ranked summaries to be more faithful. Specifically, of the 19 trials (other than the three \textit{catch trials}) where all three subjects agreed on which summary was more faithful, in 12 trials, it was the re-ranked summary (as in Table~\ref{tab:ex-all-agree}, Article 49), while in only 7 was it the original summary
(as in Table~\ref{tab:ex-all-agree}, Article 4).
In all of these cases, the authors agreed with the subjects.
Note that no information can be gleaned from those trials in which two of three subjects agreed, since in half of them (9), they agreed on the re-ranked summary, and in the other half, they agreed on the original (9). 

Finally, the reader may recall that we asked subjects after they selected a summary, whether they considered one summary to be more faithful than the other, or whether both summaries were equally faithful (or  equally unfaithful). In 21 trials, at least two subjects indicated that both summaries were equally unfaithful, even if they indicated that they felt
one summary was more faithful than the other. Often, it was because its quantity entities were closer to those in the text. For example, Table~\ref{tab:ex-all-agree}, Article 24 shows that subjects felt the original summary was more faithful since its quantity term (6\%) was closer
to the 5\% that was in the original text, while 
Table~\ref{tab:ex-all-agree}, Article 83 shows them to feel that ``by 9.5\%'' is closer to the original text than ``by 9\%'', even though the quantity in the original text is ``to 9.5\%''. In over half these trials (13/21), at least two subjects felt that the up-ranked summaries were more faithful.


\section{Conclusions}
In this paper, we addressed the problem of hallucinated quantities in summaries generated by abstractive summarization systems. We introduced \textsc{Herman}, a novel approach to recognize and verify quantities in these summaries. Experimental results demonstrate that up-ranked summaries have a higher ROUGE Precision and F$_1$ than original summaries produced by a summarization system, indicating our approach reduces hallucinated quantities while still encourage the inclusion of quantity entities. Through human evaluation, we showed that summaries up-ranked by our proposed model are felt to be more faithful than the summaries directly generated by a summarization system.

We also discovered that simple re-ranking strategies, such as the selection of the shortest summary from the beam search, can yield strong performance, if one doesn't care whether a summary communicates specific quantities. We also found that our approach was limited by its use of the XSum dataset, where factual information in the summary sometimes cannot be verified using the article due to the fact that the summary is simply the first sentence of the original article. In the future, we would like to explore the option of incorporating the verification model into training and inference to improve factual correctness of generated summaries.

\section*{Acknowledgments}

We would like to thank the anonymous reviewers, Ronald A. Cardenas, and Shashi Narayan for their helpful feedback. We also would like to thank Ronald A. Cardenas, Arlene Casey, Christian Hardmeier, and Javad Hosseini for participating in our human evaluation.

\bibliography{anthology,emnlp2020}

\begin{thebibliography}{40}
\expandafter\ifx\csname natexlab\endcsname\relax\def\natexlab#1{#1}\fi

\bibitem[{Bahdanau et~al.(2015{\natexlab{a}})Bahdanau, Cho, and
  Bengio}]{DBLP:journals/corr/BahdanauCB14}
Dzmitry Bahdanau, Kyunghyun Cho, and Yoshua Bengio. 2015{\natexlab{a}}.
\newblock \href {http://arxiv.org/abs/1409.0473} {Neural machine translation by
  jointly learning to align and translate}.
\newblock In \emph{3rd International Conference on Learning Representations,
  {ICLR} 2015, San Diego, CA, USA, May 7-9, 2015, Conference Track
  Proceedings}.

\bibitem[{Bahdanau et~al.(2015{\natexlab{b}})Bahdanau, Cho, and
  Bengio}]{BahdanauCB14}
Dzmitry Bahdanau, Kyunghyun Cho, and Yoshua Bengio. 2015{\natexlab{b}}.
\newblock \href {http://arxiv.org/abs/1409.0473} {Neural machine translation by
  jointly learning to align and translate}.
\newblock In \emph{3rd International Conference on Learning Representations,
  {ICLR} 2015, San Diego, CA, USA, May 7-9, 2015, Conference Track
  Proceedings}.

\bibitem[{Banko et~al.(2007)Banko, Cafarella, Soderland, Broadhead, and
  Etzioni}]{DBLP:conf/ijcai/BankoCSBE07}
Michele Banko, Michael~J. Cafarella, Stephen Soderland, Matthew Broadhead, and
  Oren Etzioni. 2007.
\newblock \href {http://ijcai.org/Proceedings/07/Papers/429.pdf} {Open
  information extraction from the web}.
\newblock In \emph{{IJCAI} 2007, Proceedings of the 20th International Joint
  Conference on Artificial Intelligence, Hyderabad, India, January 6-12, 2007},
  pages 2670--2676.

\bibitem[{Bowman et~al.(2015)Bowman, Angeli, Potts, and
  Manning}]{bowman-etal-2015-large}
Samuel~R. Bowman, Gabor Angeli, Christopher Potts, and Christopher~D. Manning.
  2015.
\newblock \href {https://doi.org/10.18653/v1/D15-1075} {A large annotated
  corpus for learning natural language inference}.
\newblock In \emph{Proceedings of the 2015 Conference on Empirical Methods in
  Natural Language Processing}, pages 632--642, Lisbon, Portugal. Association
  for Computational Linguistics.

\bibitem[{Cao et~al.(2018)Cao, Wei, Li, and Li}]{DBLP:conf/aaai/CaoWLL18}
Ziqiang Cao, Furu Wei, Wenjie Li, and Sujian Li. 2018.
\newblock \href
  {https://www.aaai.org/ocs/index.php/AAAI/AAAI18/paper/view/16121} {Faithful
  to the original: Fact aware neural abstractive summarization}.
\newblock In \emph{Proceedings of the Thirty-Second {AAAI} Conference on
  Artificial Intelligence, (AAAI-18), the 30th innovative Applications of
  Artificial Intelligence (IAAI-18), and the 8th {AAAI} Symposium on
  Educational Advances in Artificial Intelligence (EAAI-18), New Orleans,
  Louisiana, USA, February 2-7, 2018}, pages 4784--4791. {AAAI} Press.

\bibitem[{Clarke and Lapata(2008)}]{lapata-2005}
James Clarke and Mirella Lapata. 2008.
\newblock Global inference for sentence compression an integer linear
  programming approach.
\newblock \emph{J. Artif. Int. Res.}, 31(1):399–429.

\bibitem[{Devlin et~al.(2019)Devlin, Chang, Lee, and
  Toutanova}]{devlin-etal-2019-bert}
Jacob Devlin, Ming-Wei Chang, Kenton Lee, and Kristina Toutanova. 2019.
\newblock \href {https://doi.org/10.18653/v1/N19-1423} {{BERT}: Pre-training of
  deep bidirectional transformers for language understanding}.
\newblock In \emph{Proceedings of the 2019 Conference of the North {A}merican
  Chapter of the Association for Computational Linguistics: Human Language
  Technologies, Volume 1 (Long and Short Papers)}, pages 4171--4186,
  Minneapolis, Minnesota. Association for Computational Linguistics.

\bibitem[{Falke et~al.(2019)Falke, Ribeiro, Utama, Dagan, and
  Gurevych}]{falke-etal-2019-ranking}
Tobias Falke, Leonardo F.~R. Ribeiro, Prasetya~Ajie Utama, Ido Dagan, and Iryna
  Gurevych. 2019.
\newblock \href {https://doi.org/10.18653/v1/P19-1213} {Ranking generated
  summaries by correctness: An interesting but challenging application for
  natural language inference}.
\newblock In \emph{Proceedings of the 57th Annual Meeting of the Association
  for Computational Linguistics}, pages 2214--2220, Florence, Italy.
  Association for Computational Linguistics.

\bibitem[{Gardner et~al.(2018)Gardner, Grus, Neumann, Tafjord, Dasigi, Liu,
  Peters, Schmitz, and Zettlemoyer}]{DBLP:journals/corr/abs-1803-07640}
Matt Gardner, Joel Grus, Mark Neumann, Oyvind Tafjord, Pradeep Dasigi,
  Nelson~F. Liu, Matthew~E. Peters, Michael Schmitz, and Luke Zettlemoyer.
  2018.
\newblock \href {http://arxiv.org/abs/1803.07640} {Allennlp: {A} deep semantic
  natural language processing platform}.
\newblock \emph{CoRR}, abs/1803.07640.

\bibitem[{Goodrich et~al.(2019)Goodrich, Rao, Liu, and
  Saleh}]{goodrich-2019-assess}
Ben Goodrich, Vinay Rao, Peter~J. Liu, and Mohammad Saleh. 2019.
\newblock \href {https://doi.org/10.1145/3292500.3330955} {Assessing the
  factual accuracy of generated text}.
\newblock In \emph{Proceedings of the 25th ACM SIGKDD International Conference
  on Knowledge Discovery \& Data Mining}, KDD ’19, page 166–175, New York,
  NY, USA. Association for Computing Machinery.

\bibitem[{Grusky et~al.(2018)Grusky, Naaman, and
  Artzi}]{DBLP:conf/naacl/GruskyNA18}
Max Grusky, Mor Naaman, and Yoav Artzi. 2018.
\newblock \href {https://doi.org/10.18653/v1/n18-1065} {Newsroom: {A} dataset
  of 1.3 million summaries with diverse extractive strategies}.
\newblock In \emph{Proceedings of the 2018 Conference of the North American
  Chapter of the Association for Computational Linguistics: Human Language
  Technologies, {NAACL-HLT} 2018, New Orleans, Louisiana, USA, June 1-6, 2018,
  Volume 1 (Long Papers)}, pages 708--719. Association for Computational
  Linguistics.

\bibitem[{Hermann et~al.(2015)Hermann, Kocisk{\'{y}}, Grefenstette, Espeholt,
  Kay, Suleyman, and Blunsom}]{DBLP:conf/nips/HermannKGEKSB15}
Karl~Moritz Hermann, Tom{\'{a}}s Kocisk{\'{y}}, Edward Grefenstette, Lasse
  Espeholt, Will Kay, Mustafa Suleyman, and Phil Blunsom. 2015.
\newblock \href
  {http://papers.nips.cc/paper/5945-teaching-machines-to-read-and-comprehend}
  {Teaching machines to read and comprehend}.
\newblock In \emph{Advances in Neural Information Processing Systems 28: Annual
  Conference on Neural Information Processing Systems 2015, December 7-12,
  2015, Montreal, Quebec, Canada}, pages 1693--1701.

\bibitem[{Hochreiter and Schmidhuber(1997)}]{hochreiter1997long}
Sepp Hochreiter and J\"{u}rgen Schmidhuber. 1997.
\newblock \href {https://doi.org/10.1162/neco.1997.9.8.1735} {Long short-term
  memory}.
\newblock \emph{Neural Comput.}, 9(8):1735–1780.

\bibitem[{Honnibal and Montani(2017)}]{spacy2}
Matthew Honnibal and Ines Montani. 2017.
\newblock {spaCy 2}: Natural language understanding with {B}loom embeddings,
  convolutional neural networks and incremental parsing.
\newblock \url{http://spacy.io}.

\bibitem[{Huang et~al.(2015)Huang, Xu, and Yu}]{Huang2015BidirectionalLM}
Zhiheng Huang, Wei Xu, and Kai Yu. 2015.
\newblock Bidirectional lstm-crf models for sequence tagging.
\newblock \emph{ArXiv}, abs/1508.01991.

\bibitem[{Kingma and Ba(2015)}]{KingmaB14}
Diederik~P. Kingma and Jimmy Ba. 2015.
\newblock \href {http://arxiv.org/abs/1412.6980} {Adam: {A} method for
  stochastic optimization}.
\newblock In \emph{3rd International Conference on Learning Representations,
  {ICLR} 2015, San Diego, CA, USA, May 7-9, 2015, Conference Track
  Proceedings}.

\bibitem[{Kryscinski et~al.(2019{\natexlab{a}})Kryscinski, Keskar, McCann,
  Xiong, and Socher}]{kryscinski-etal-2019-neural}
Wojciech Kryscinski, Nitish~Shirish Keskar, Bryan McCann, Caiming Xiong, and
  Richard Socher. 2019{\natexlab{a}}.
\newblock \href {https://doi.org/10.18653/v1/D19-1051} {Neural text
  summarization: A critical evaluation}.
\newblock In \emph{Proceedings of the 2019 Conference on Empirical Methods in
  Natural Language Processing and the 9th International Joint Conference on
  Natural Language Processing (EMNLP-IJCNLP)}, pages 540--551, Hong Kong,
  China. Association for Computational Linguistics.

\bibitem[{Kryscinski et~al.(2019{\natexlab{b}})Kryscinski, McCann, Xiong, and
  Socher}]{DBLP:journals/corr/abs-1910-12840}
Wojciech Kryscinski, Bryan McCann, Caiming Xiong, and Richard Socher.
  2019{\natexlab{b}}.
\newblock \href {http://arxiv.org/abs/1910.12840} {Evaluating the factual
  consistency of abstractive text summarization}.
\newblock \emph{CoRR}, abs/1910.12840.

\bibitem[{Lafferty et~al.(2001)Lafferty, McCallum, and Pereira}]{LaffertyMP01}
John~D. Lafferty, Andrew McCallum, and Fernando C.~N. Pereira. 2001.
\newblock Conditional random fields: Probabilistic models for segmenting and
  labeling sequence data.
\newblock In \emph{Proceedings of the Eighteenth International Conference on
  Machine Learning {(ICML} 2001), Williams College, Williamstown, MA, USA, June
  28 - July 1, 2001}, pages 282--289. Morgan Kaufmann.

\bibitem[{Lewis et~al.(2019)Lewis, Liu, Goyal, Ghazvininejad, Mohamed, Levy,
  Stoyanov, and Zettlemoyer}]{bart-abs-1910-13461}
Mike Lewis, Yinhan Liu, Naman Goyal, Marjan Ghazvininejad, Abdelrahman Mohamed,
  Omer Levy, Veselin Stoyanov, and Luke Zettlemoyer. 2019.
\newblock \href {http://arxiv.org/abs/1910.13461} {{BART:} denoising
  sequence-to-sequence pre-training for natural language generation,
  translation, and comprehension}.
\newblock \emph{CoRR}, abs/1910.13461.

\bibitem[{Lin(2004)}]{lin-2004-rouge}
Chin-Yew Lin. 2004.
\newblock \href {https://www.aclweb.org/anthology/W04-1013} {{ROUGE}: A package
  for automatic evaluation of summaries}.
\newblock In \emph{Text Summarization Branches Out}, pages 74--81, Barcelona,
  Spain. Association for Computational Linguistics.

\bibitem[{Lin and Bilmes(2011)}]{lin-bilmes-2011-class}
Hui Lin and Jeff Bilmes. 2011.
\newblock \href {https://www.aclweb.org/anthology/P11-1052} {A class of
  submodular functions for document summarization}.
\newblock In \emph{Proceedings of the 49th Annual Meeting of the Association
  for Computational Linguistics: Human Language Technologies}, pages 510--520,
  Portland, Oregon, USA. Association for Computational Linguistics.

\bibitem[{Liu and Lapata(2019)}]{liu-lapata-2019-text}
Yang Liu and Mirella Lapata. 2019.
\newblock \href {https://doi.org/10.18653/v1/D19-1387} {Text summarization with
  pretrained encoders}.
\newblock In \emph{Proceedings of the 2019 Conference on Empirical Methods in
  Natural Language Processing and the 9th International Joint Conference on
  Natural Language Processing (EMNLP-IJCNLP)}, pages 3730--3740, Hong Kong,
  China. Association for Computational Linguistics.

\bibitem[{Louis and Nenkova(2011)}]{louis-nenkova-2011-text}
Annie Louis and Ani Nenkova. 2011.
\newblock \href {https://www.aclweb.org/anthology/W11-1605} {Text specificity
  and impact on quality of news summaries}.
\newblock In \emph{Proceedings of the Workshop on Monolingual Text-To-Text
  Generation}, pages 34--42, Portland, Oregon. Association for Computational
  Linguistics.

\bibitem[{Maynez et~al.(2020)Maynez, Narayan, Bohnet, and
  McDonald}]{DBLP:journals/corr/abs-2005-00661}
Joshua Maynez, Shashi Narayan, Bernd Bohnet, and Ryan~T. McDonald. 2020.
\newblock \href {http://arxiv.org/abs/2005.00661} {On faithfulness and
  factuality in abstractive summarization}.
\newblock \emph{CoRR}, abs/2005.00661.

\bibitem[{Nallapati et~al.(2017)Nallapati, Zhai, and
  Zhou}]{DBLP:conf/aaai/NallapatiZZ17}
Ramesh Nallapati, Feifei Zhai, and Bowen Zhou. 2017.
\newblock \href {http://aaai.org/ocs/index.php/AAAI/AAAI17/paper/view/14636}
  {Summarunner: {A} recurrent neural network based sequence model for
  extractive summarization of documents}.
\newblock In \emph{Proceedings of the Thirty-First {AAAI} Conference on
  Artificial Intelligence, February 4-9, 2017, San Francisco, California,
  {USA}}, pages 3075--3081. {AAAI} Press.

\bibitem[{Narayan et~al.(2018{\natexlab{a}})Narayan, Cohen, and
  Lapata}]{narayan-etal-2018-dont}
Shashi Narayan, Shay~B. Cohen, and Mirella Lapata. 2018{\natexlab{a}}.
\newblock \href {https://doi.org/10.18653/v1/D18-1206} {Don{'}t give me the
  details, just the summary! topic-aware convolutional neural networks for
  extreme summarization}.
\newblock In \emph{Proceedings of the 2018 Conference on Empirical Methods in
  Natural Language Processing}, pages 1797--1807, Brussels, Belgium.
  Association for Computational Linguistics.

\bibitem[{Narayan et~al.(2018{\natexlab{b}})Narayan, Cohen, and
  Lapata}]{narayan-etal-2018-ranking}
Shashi Narayan, Shay~B. Cohen, and Mirella Lapata. 2018{\natexlab{b}}.
\newblock \href {https://doi.org/10.18653/v1/N18-1158} {Ranking sentences for
  extractive summarization with reinforcement learning}.
\newblock In \emph{Proceedings of the 2018 Conference of the North {A}merican
  Chapter of the Association for Computational Linguistics: Human Language
  Technologies, Volume 1 (Long Papers)}, pages 1747--1759, New Orleans,
  Louisiana. Association for Computational Linguistics.

\bibitem[{Paszke et~al.(2019)Paszke, Gross, Massa, Lerer, Bradbury, Chanan,
  Killeen, Lin, Gimelshein, Antiga, Desmaison, Kopf, Yang, DeVito, Raison,
  Tejani, Chilamkurthy, Steiner, Fang, Bai, and Chintala}]{NEURIPS2019_9015}
Adam Paszke, Sam Gross, Francisco Massa, Adam Lerer, James Bradbury, Gregory
  Chanan, Trevor Killeen, Zeming Lin, Natalia Gimelshein, Luca Antiga, Alban
  Desmaison, Andreas Kopf, Edward Yang, Zachary DeVito, Martin Raison, Alykhan
  Tejani, Sasank Chilamkurthy, Benoit Steiner, Lu~Fang, Junjie Bai, and Soumith
  Chintala. 2019.
\newblock \href
  {http://papers.neurips.cc/paper/9015-pytorch-an-imperative-style-high-performance-deep-learning-library.pdf}
  {Pytorch: An imperative style, high-performance deep learning library}.
\newblock In H.~Wallach, H.~Larochelle, A.~Beygelzimer, F.~d~Alch\'{e}-Buc,
  E.~Fox, and R.~Garnett, editors, \emph{Advances in Neural Information
  Processing Systems 32}, pages 8024--8035. Curran Associates, Inc.

\bibitem[{Pennington et~al.(2014)Pennington, Socher, and
  Manning}]{pennington2014glove}
Jeffrey Pennington, Richard Socher, and Christopher~D. Manning. 2014.
\newblock \href {http://www.aclweb.org/anthology/D14-1162} {Glove: Global
  vectors for word representation}.
\newblock In \emph{Empirical Methods in Natural Language Processing (EMNLP)},
  pages 1532--1543.

\bibitem[{Ramshaw and Marcus(1999)}]{ramshaw1999text}
Lance~A Ramshaw and Mitchell~P Marcus. 1999.
\newblock Text chunking using transformation-based learning.
\newblock In \emph{Natural language processing using very large corpora}, pages
  157--176. Springer.

\bibitem[{Rush et~al.(2015)Rush, Chopra, and Weston}]{rush-etal-2015-neural}
Alexander~M. Rush, Sumit Chopra, and Jason Weston. 2015.
\newblock \href {https://doi.org/10.18653/v1/D15-1044} {A neural attention
  model for abstractive sentence summarization}.
\newblock In \emph{Proceedings of the 2015 Conference on Empirical Methods in
  Natural Language Processing}, pages 379--389, Lisbon, Portugal. Association
  for Computational Linguistics.

\bibitem[{Sandhaus(2008)}]{sandhaus2008new}
Evan Sandhaus. 2008.
\newblock The new york times annotated corpus.
\newblock \emph{Linguistic Data Consortium, Philadelphia}, 6(12):e26752.

\bibitem[{See et~al.(2017)See, Liu, and Manning}]{see-etal-2017-get}
Abigail See, Peter~J. Liu, and Christopher~D. Manning. 2017.
\newblock \href {https://doi.org/10.18653/v1/P17-1099} {Get to the point:
  Summarization with pointer-generator networks}.
\newblock In \emph{Proceedings of the 55th Annual Meeting of the Association
  for Computational Linguistics (Volume 1: Long Papers)}, pages 1073--1083,
  Vancouver, Canada. Association for Computational Linguistics.

\bibitem[{Sutskever et~al.(2014)Sutskever, Vinyals, and
  Le}]{DBLP:conf/nips/SutskeverVL14}
Ilya Sutskever, Oriol Vinyals, and Quoc~V. Le. 2014.
\newblock \href
  {http://papers.nips.cc/paper/5346-sequence-to-sequence-learning-with-neural-networks}
  {Sequence to sequence learning with neural networks}.
\newblock In \emph{Advances in Neural Information Processing Systems 27: Annual
  Conference on Neural Information Processing Systems 2014, December 8-13 2014,
  Montreal, Quebec, Canada}, pages 3104--3112.

\bibitem[{Vaswani et~al.(2017)Vaswani, Shazeer, Parmar, Uszkoreit, Jones,
  Gomez, Kaiser, and Polosukhin}]{NIPS2017_7181}
Ashish Vaswani, Noam Shazeer, Niki Parmar, Jakob Uszkoreit, Llion Jones,
  Aidan~N Gomez, \L~ukasz Kaiser, and Illia Polosukhin. 2017.
\newblock \href
  {http://papers.nips.cc/paper/7181-attention-is-all-you-need.pdf} {Attention
  is all you need}.
\newblock In I.~Guyon, U.~V. Luxburg, S.~Bengio, H.~Wallach, R.~Fergus,
  S.~Vishwanathan, and R.~Garnett, editors, \emph{Advances in Neural
  Information Processing Systems 30}, pages 5998--6008. Curran Associates, Inc.

\bibitem[{Vinyals et~al.(2015)Vinyals, Fortunato, and
  Jaitly}]{DBLP:conf/nips/VinyalsFJ15}
Oriol Vinyals, Meire Fortunato, and Navdeep Jaitly. 2015.
\newblock \href {http://papers.nips.cc/paper/5866-pointer-networks} {Pointer
  networks}.
\newblock In \emph{Advances in Neural Information Processing Systems 28: Annual
  Conference on Neural Information Processing Systems 2015, December 7-12,
  2015, Montreal, Quebec, Canada}, pages 2692--2700.

\bibitem[{Wang et~al.(2019)Wang, Quan, and Wang}]{wang-etal-2019-biset}
Kai Wang, Xiaojun Quan, and Rui Wang. 2019.
\newblock \href {https://doi.org/10.18653/v1/P19-1207} {{B}i{SET}:
  Bi-directional selective encoding with template for abstractive
  summarization}.
\newblock In \emph{Proceedings of the 57th Annual Meeting of the Association
  for Computational Linguistics}, pages 2153--2162, Florence, Italy.
  Association for Computational Linguistics.

\bibitem[{Wolf et~al.(2019)Wolf, Debut, Sanh, Chaumond, Delangue, Moi, Cistac,
  Rault, Louf, Funtowicz, and Brew}]{DBLP:journals/corr/abs-1910-03771}
Thomas Wolf, Lysandre Debut, Victor Sanh, Julien Chaumond, Clement Delangue,
  Anthony Moi, Pierric Cistac, Tim Rault, R{\'{e}}mi Louf, Morgan Funtowicz,
  and Jamie Brew. 2019.
\newblock \href {http://arxiv.org/abs/1910.03771} {Huggingface's transformers:
  State-of-the-art natural language processing}.
\newblock \emph{CoRR}, abs/1910.03771.

\bibitem[{Zhang et~al.(2019)Zhang, Merck, Tsai, Manning, and
  Langlotz}]{DBLP:journals/corr/abs-1911-02541}
Yuhao Zhang, Derek Merck, Emily~Bao Tsai, Christopher~D. Manning, and Curtis~P.
  Langlotz. 2019.
\newblock \href {http://arxiv.org/abs/1911.02541} {Optimizing the factual
  correctness of a summary: {A} study of summarizing radiology reports}.
\newblock \emph{CoRR}, abs/1911.02541.

\end{thebibliography}
\bibliographystyle{acl_natbib}

\clearpage
\appendix

\section{Supplementary Material}
\label{sec:appendix}
This appendix provides details of training for our \textsc{Herman} model, in addition to experiment settings mention in Section \ref{sec:experiment}. Our \textsc{Herman} model has 19,424,661 parameters in total. On a single GeForce GTX 1080 Ti GPU, with batch size of 32 and using GloVe vectors (6B tokens, \textit{uncased}) for word embeddings initialization, our \textsc{Herman} model need approximately 1 hour to train one epoch. With the same GPU and batch size, \textsc{Herman} model with pre-trained \textit{base-uncased} BERT for word embedding initialization requires 3 hours to train one epoch. We use the Huggingface Transformers library \citep{DBLP:journals/corr/abs-1910-03771} for BERT word embedding initialization. 

As mentioned in Section \ref{sec:experiment}, we are using three summarization systems, \textsc{TConvS2S}, \textsc{BertSum}, and BART for getting the beam of summaries to be re-ranked by \textsc{Herman}. For \textsc{BertSum}, we use the abstractive model variant \textsc{BertSumExtAbs} which gets the best performance for XSum. We use the same beam size as reported by the authors. For \textsc{TConvS2S}, \textsc{BertSum}, and BART, beam size used are 10, 5, and 6, respectively. We did hyperparameter search for $\alpha$ which indicates weight balance between $\mathcal{L}_\mathit{Y}$ and $\mathcal{L}_\mathit{z}$. Our search space is $[0,1]$, with three configurations, $\alpha=0.33$, $\alpha=0.5$, and $\alpha=0.66$. We choose the best configuration, $\alpha=0.66$, based on the loss on the validation set. 

\end{document}